\PassOptionsToPackage{prologue,dvipsnames}{xcolor}
\documentclass{article} 
\usepackage{iclr2025_conference,times}

\usepackage{amsmath,amsfonts,bm}

\def\eqref#1{equation~\ref{#1}}

\def\1{\bm{1}}

\DeclareMathAlphabet{\mathsfit}{\encodingdefault}{\sfdefault}{m}{sl}
\SetMathAlphabet{\mathsfit}{bold}{\encodingdefault}{\sfdefault}{bx}{n}

\usepackage[dvipsnames]{xcolor}
\usepackage{hyperref}
\hypersetup{colorlinks,
           linkcolor=NavyBlue,
           urlcolor=NavyBlue,
           citecolor=NavyBlue}

\usepackage{url}
\usepackage{graphicx}
\usepackage{tabu}
\usepackage{booktabs}
\usepackage{multirow}
\usepackage{colortbl}
\usepackage{subcaption}
\usepackage[font=small]{caption}
\usepackage{wrapfig}
\usepackage{arydshln} 
\usepackage{makecell}
\usepackage{stfloats}
\usepackage{array}
\usepackage[hang,flushmargin]{footmisc}
\usepackage{wasysym}
\usepackage{listings}
\usepackage{cleveref}
\usepackage[dvipsnames]{xcolor}

\definecolor{mygreen}{RGB}{29, 177, 0}
\definecolor{darkgreen}{rgb}{0.0, 0.5, 0.0}

\def\ourmethod{ReKV}

\newcommand{\plusvalue}[1]{\hspace{0.2em}\textcolor{darkgreen}{\textbf{\scriptsize{(+#1)}}}}
\newcommand{\minusvalue}[1]{\hspace{0.2em}\textcolor{red}{\textbf{\scriptsize{(-#1)}}}}
\newcommand{\retrieve}[2]{#1 FPS $\to$ #2 Frames}

\title{Streaming Video Question-Answering with \\In-context Video KV-Cache Retrieval}

\author{
Shangzhe Di$^{1,2}$
\quad
Zhelun Yu$^{2}$
\quad
Guanghao Zhang$^{2}$
\quad
Haoyuan Li$^{2}$
\quad
Tao Zhong$^{2}$
\\
\textbf{
Hao Cheng$^{2}$
\quad
Bolin Li$^{2}$
\quad
Wanggui He$^{2}$
\quad
Fangxun Shu$^{2}$
\quad
Hao Jiang$^{2}$
}
\vspace{3pt}
\\
$^{1}$Shanghai Jiao Tong University \quad $^{2}$Alibaba Group
\\
\texttt{dishangzhe@sjtu.edu.cn}
}

\iclrfinalcopy 
\begin{document}

\maketitle

\begin{abstract}

We propose \textbf{\ourmethod}, a novel training-free approach that enables efficient streaming video question-answering (StreamingVQA),
by seamlessly integrating with existing Video Large Language Models (Video-LLMs).
Traditional VideoQA systems struggle with long videos, as they must process entire videos before responding to queries, and repeat this process for each new question. In contrast, our approach analyzes long videos in a streaming manner, allowing for prompt responses as soon as user queries are received. Building on a common Video-LLM, we first incorporate a sliding-window attention mechanism, ensuring that input frames attend to a limited number of preceding frames, thereby reducing computational overhead.
To prevent information loss, we store processed video key-value caches (KV-Caches) in RAM and disk, reloading them into GPU memory as needed. Additionally, we introduce a retrieval method that leverages an external retriever or the parameters within Video-LLMs to retrieve only query-relevant KV-Caches, ensuring both efficiency and accuracy in question answering. \ourmethod~enables the separation of video encoding and question-answering across different processes and GPUs, significantly enhancing the efficiency of StreamingVQA. Through comprehensive experimentation, we validate the efficacy and practicality of our approach, which significantly boosts efficiency and enhances applicability over existing VideoQA models.
    
\end{abstract}
\section{Introduction}
\label{sec:introduction}

In the literature, video understanding tasks, 
such as action recognition~\citep{activitynet,something,kinetics}, 
visual object tracking~\citep{got_10k,trackingnet}, 
and video question-answering~\citep{msrvtt_qa,jang2017tgifqa,next_qa,li2024mvbench}, 
have primarily focused on short clips lasting from a few seconds to minutes. 
However, as vision models increasingly find applications in real-world scenarios like robotics, surveillance, and live broadcasts, the research in the vision community has gradually shifted towards understanding continuous video streams, where long-term contexts and real-time interaction are crucial.

\begin{figure}[t]
\centerline{\includegraphics[width=\linewidth]{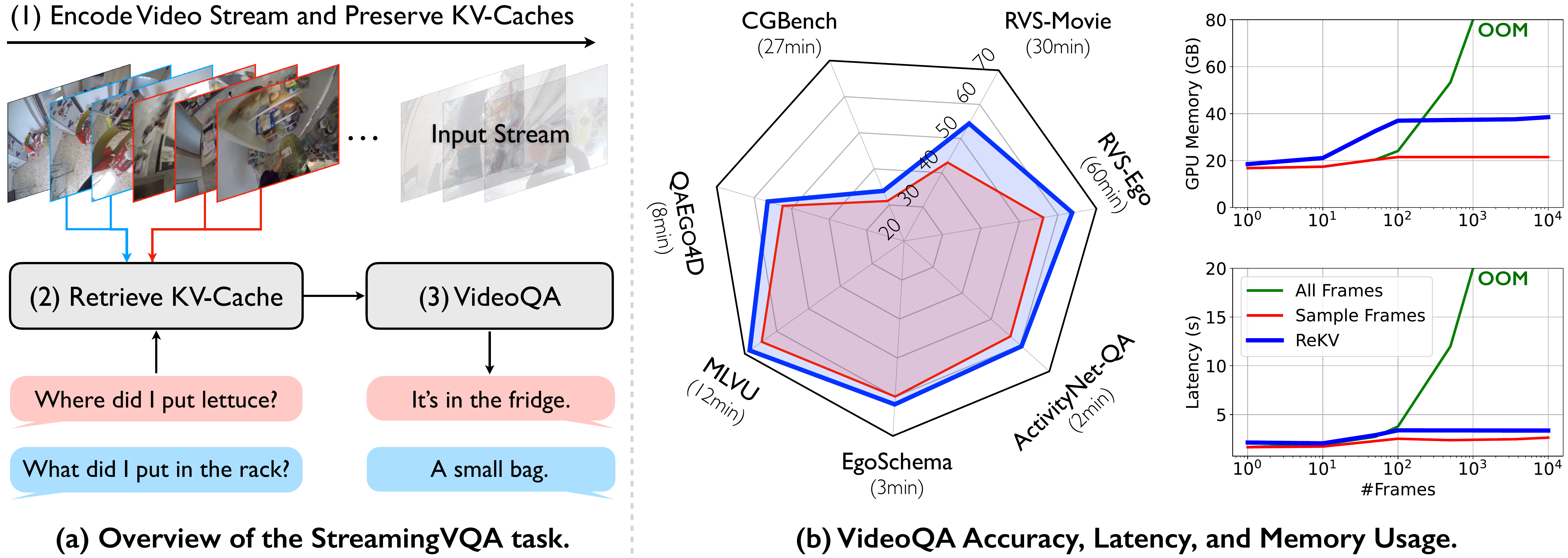}}
\caption{
\textbf{Overview of the StreamingVQA task and our proposed \ourmethod.}
(a) StreamingVQA requires a model to continuously process video streams and answer questions about previously viewed content at any moment. 
(b) We propose \ourmethod~to enhance efficiency and accuracy in StreamingVQA.
Tested with \texttt{LLaVA-OV-7B} on an H800 (80GB) GPU, \ourmethod~maintains stable latency and GPU memory usage, preventing out-of-memory (OOM) errors as frames increase. It also improves the accuracy on seven long-form VideoQA benchmarks compared to the uniform sampling baseline. Further details are provided in Section~\ref{sec:experiments}.
}
\label{fig:teaser} 
\end{figure}

In this paper, we consider the problem of \textbf{streaming video question-answering (StreamingVQA)}. As shown in Figure~\ref{fig:teaser}(a),
it involves continuously processing long video streams and promptly responding to queries about the visual content at any moment. It can be treated as a generalization of the standard offline VideoQA, where the model processes the entire video and all questions simultaneously. By definition, such task of StreamingVQA presents three core challenges: 
(i) \textbf{Efficient Video Encoding:}
Unlike traditional offline VideoQA, where models have access to the entire video clip, StreamingVQA demands real-time analysis of continuous streams. Models must efficiently process incoming frames without access to future frames or frequent revisiting of distant past frames. (ii) \textbf{Video Context Preservation:} 
To accurately answer questions posed later in the stream, models must preserve relevant information from earlier frames, making long-term context retention a key challenge. 
(iii) \textbf{Real-Time Response:} 
The model must provide accurate answers with minimal delay, requiring efficient retrieval of video context and rapid question-answering.

Current Video-LLMs often struggle to encode long video streams due to the large volume of video tokens, forcing most models to process only a sparse subset of frames~\citep{video_chatgpt, llava_next_video, llava_onevision}. 
This results in limited video lengths or a significant loss of fine-grained visual information. While techniques like average pooling~\citep{llamavid} and memory compression~\citep{wu2022memvit,wang2023memory,he2024ma,flashvstream,videostreaming} reduce token volume, they come at the cost of losing details, particularly in temporal and lower-level visual features that are essential for complex question answering.

To address the challenges, we propose \textbf{\ourmethod} (\textbf{Re}trieve In-context Video \textbf{KV}-Cache), a framework that seamlessly integrates with existing Video-LLMs~\citep{video_chatgpt, llava_next_video, llava_onevision} without additional training. 
Our method employs two strategies for aggregating both short- and long-term temporal information. 
For \textbf{short-term temporal context}, 
the model adopts causal attention with a sliding-window mechanism~\citep{lm_infinite}, where tokens attend only to a limited set of preceding tokens during encoding.  
For \textbf{recalling long-term information}, 
we enable dynamic access to any point within the video sequence via retrieval. 
Specifically, our method retains and reuses past computations (KV-Cache) to avoid redundant processing while enhancing long-term reasoning without sacrificing detail. For extremely long videos, KV-Caches can be offloaded to RAM or disk to prevent memory overflow.

To ensure real-time and accurate responses, we retrieve a fixed number of KV-Caches relevant to the current question. This design strikes a balance between efficiency and accuracy by avoiding the need to process all past frames, while still accessing the most critical information. We experimented with two retrieval methods: one using external CLIP-like models~\citep{clip, siglip} for semantic matching, and another leveraging internal attention weights for faster, more integrated, and potentially stronger retrieval~\citep{infllm, snapkv}.

In summary, {\ourmethod}~efficiently encodes long video streams, 
preserves and retrieves in-context KV-Caches to address complex video question-answering. In addition, {\ourmethod}~separates video encoding from question-answering into distinct processes, further enhancing efficiency. As shown in Figure~\ref{fig:teaser}(b), {\ourmethod}~improves VideoQA accuracy while maintaining stable inference latency and memory usage as frames increase. 
The remainder of the paper is organized as follows: 
Section~\ref{sec:related_work} provides an overview of the relevant literature. 
Section~\ref{sec:method} formulates the StreamingVQA task and describes our proposed method in detail. In Section~\ref{sec:experiments}, we present ablation studies and comparisons to validate our approach. Consequently, our approach not only enhances accuracy on long VideoQA benchmarks, including MLVU~\citep{mlvu}, \textsc{QAEgo4D$_\texttt{MC}$}~\citep{di2023groundvqa}, EgoSchema~\citep{egoschema}, and ActivityNet-QA~\citep{activitynet_qa}, as well as StreamingVQA benchmarks~\citep{flashvstream}, but also reduces inference latency and memory usage.

\section{StreamingVQA: Task Definition and Discussion} 
\label{sec:task}

This paper considers the problem of streaming video question-answering (\textbf{StreamingVQA}), where a model continuously processes a video stream and can respond to questions about past visual content at any moment. 
In this section, we formally define the task and outline the design principles for our proposed solution.

Given a video stream $\mathcal{V}^T := [v_1, v_2, ..., v_T]$ consisting of $T$ frames and a set of $N$ questions $\mathcal{Q} := \{q_1, q_2, \dots, q_N\}$,
StreamingVQA aims to answer a question $q_i$ at any time step $t~(1 \le t \le T)$, using only the frames seen up to that point, $\mathcal{V}^t := [v_1, v_2, ..., v_t]$.

\noindent\textbf{Discussion-I: StreamingVQA {\em vs.}~OfflineVQA.} StreamingVQA involves continuously analyzing an incoming video stream and answering questions based on the observed visual content at any moment. 
In contrast, conventional video question-answering models~\citep{frozen_bilm,video_chatgpt,llava_next_video,llava_onevision} operate in an offline mode, referred to as OfflineVQA.
The two paradigms differ in that: 
1) StreamingVQA processes a continuous video stream, while OfflineVQA handles a predefined video input, and 
2) StreamingVQA allows questions to be asked at any point during the stream, whereas OfflineVQA processes questions only after the entire video has been viewed.
Notably, OfflineVQA can be considered a special case of StreamingVQA, where all questions are posed after the video is fully processed.

Conventional approaches typically employ a visual encoder~\citep{clip,siglip,fang2023eva} and a projection module~\citep{llava_next_video,blip2} to process video frames~($\mathcal{V}^t$).
The output is concatenated with the tokenized question to form a sequence $[\mathcal{V}_t, q_i]$~\footnote{We maintain the original notation for simplicity.}, which is then passed to an LLMs to predict an answer.
However, this approach is impractical due to the high computational cost associated with processing a large number of frames ($T$).

A common workaround is sparse frame sampling~\citep{video_chatgpt, llava_next_video, llava_onevision}, 
but this introduces new problems:
(i) loss of critical visual information, leading to incomplete or inaccurate responses, and
(ii) the need to reprocess frames for different questions, since questions asked at different time points require distinct frame samples. This becomes increasingly inefficient as $T$ and $N$ grow.

Given these challenges, current OfflineVQA methods fall short when applied to StreamingVQA scenarios. Therefore, designing a new approach optimized for StreamingVQA is crucial to handling video streams more efficiently, enabling real-time question answering and unlocking more interactive video analysis applications.

\noindent\textbf{Discussion-II: Design Principles for Efficient StreamingVQA.} 
To tackle the aforementioned challenges, we can exploit the causal nature of the LLM decoder to avoid redundant computations and strike a balance between accuracy and speed. 
During attention calculations, causal masking prevents the model from accessing future tokens, ensuring that video tokens are encoded independently of the questions. This allows us to \textit{decouple video encoding from question-answering}.

For video encoding, we leverage the KV-Cache optimization to accelerate inference.
However, as number of frames grows large, handling the massive number of video tokens becomes increasingly inefficient and may exceed the model’s capacity~\citep{lm_infinite,streamingllm}.
To address this, we adopt a sliding-window attention mechanism~\citep{lm_infinite}, which limits the attention scope to only the most recent frames.

Regarding question-answering, Video KV-Caches are stored and can be reused as context to answer different questions. 
However, long video sequences produce a substantial amount of KV-Caches, leading to excessive GPU memory consumption, computational overhead, and unnecessary distractions if all are used.
To address this, we introduce an efficient retrieval method that selects the most relevant video key-value vectors from the video KV-Caches. These selected vectors then serve as context, enabling efficient and scalable StreamingVQA.

\section{
\ourmethod: \underline{Re}trieve In-context Video \underline{KV}-Cache
}
\label{sec:method}

This section introduces \textbf{\ourmethod}, an approach that integrates seamlessly with a Video-LLM to enable efficient StreamingVQA without requiring additional training.
Overall, \ourmethod~efficiently encodes the video stream, maintains its KV-Caches, retrieves relevant caches based on the given question, and uses them for accurate question-answering.

\begin{figure}[t]
\centerline{\includegraphics[width=.85\linewidth]{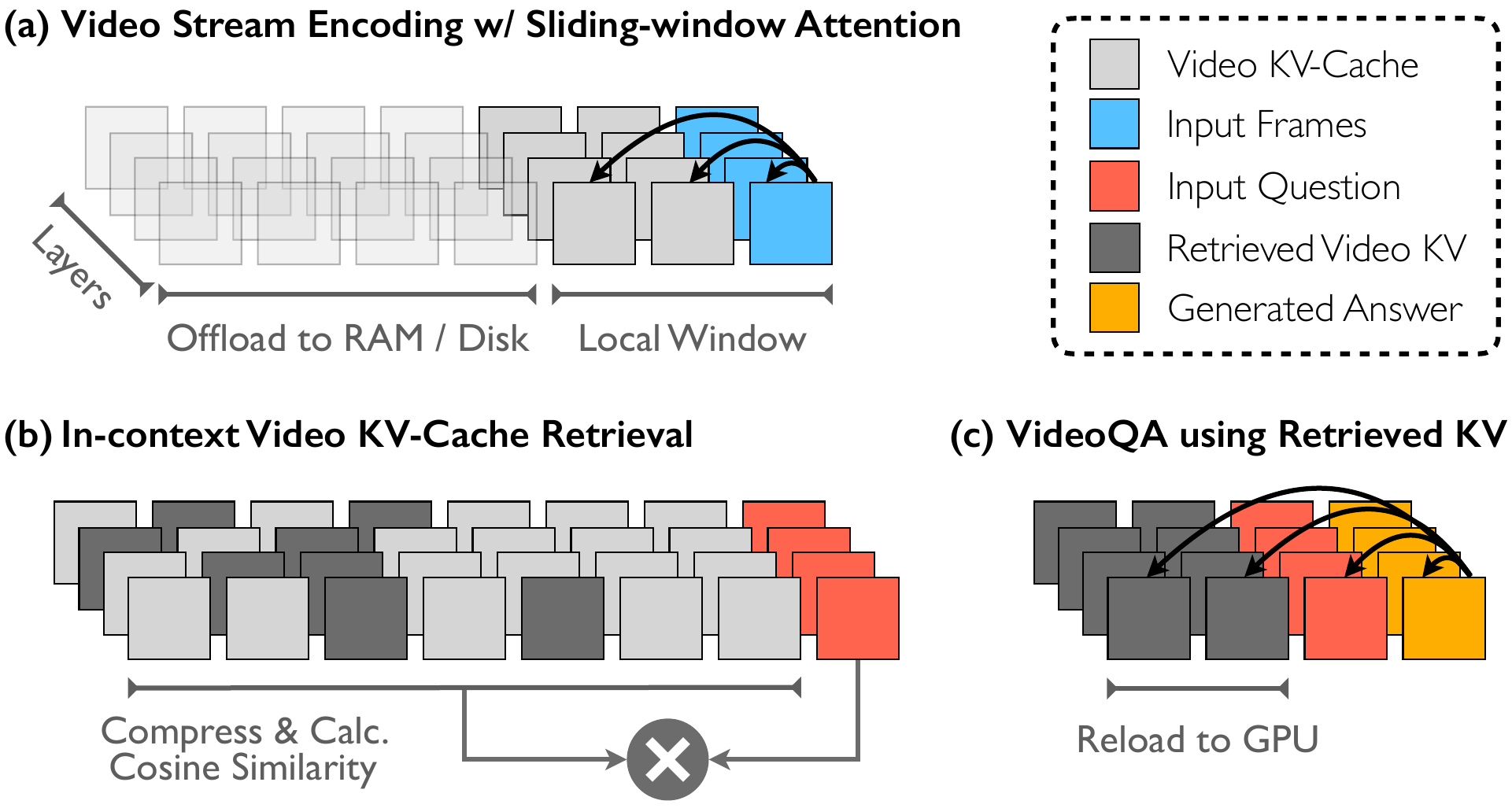}}
\caption{
\textbf{Overview of~\ourmethod.}
We modify the attention mechanism in Decoder-based Video-LLMs:
(a) The video stream is encoded with sliding-window attention (Equation~\ref{equ:video_encoding}), with out-of-window Video KV-Caches offloaded to RAM or disk.
(b) Upon receiving a question, relevant key-value vectors are retrieved based on cosine similarity, with compressed vectors to accelerate retrieval (Equation~\ref{equ:retrieval}).
(c) The retrieved key-value vectors are reloaded onto the GPU and utilized for autoregressive answer generation (Equation~\ref{equ:answer_generation}).
}
\label{fig:framework} 
\end{figure}

We aim to enable Video-LLMs, trained on limited frames, 
to perform StreamingVQA \textbf{without additional training}.
As shown in Figure~\ref{fig:framework}, 
the proposed \ourmethod~has three components: 
video stream encoding, video KV-Cache retrieval, and question-answering using the retrieved key-value vectors.

\noindent\textbf{Video Stream Encoding with Sliding-window Attention.} 
We encode the video stream $\mathcal{V}^T$ incrementally, processing it chunk by chunk.
At each step, the inputs include past key-value vectors $\mathbf{P} = \{(\mathbf{k}_j, \mathbf{v}_j)\}_{j=1}^{l_P}$ and the current tokens $\mathbf{X}=\{\mathbf{t}_{i+l_P}\}_{i=1}^{l_X}$, 
where $l_P$ denotes the lengths of past key-values, and $l_X$ refers to the chunk size.
The local key-value vectors within a window $l_L$ can thus be derived as 
$\mathbf{L} = \mathbf{P}_{[l_P - l_L + 1 : l_P]}$.
The attention calculation is then formulated as:
\begin{equation}
    \mathbf{O} = \text{Attn}\left(\mathbf{W_Q}\mathbf{X}, [\mathbf{L}_k, \mathbf{W_K}\mathbf{X}], [\mathbf{L}_v, \mathbf{W_V}\mathbf{X}]\right),
    \label{equ:video_encoding}
\end{equation}
where $\mathbf{W_Q}$, $\mathbf{W_K}$, and $\mathbf{W_V}$ are the attention layer parameters, $\mathbf{L}_k$ and $\mathbf{L}_v$ correspond to the key and value vectors in $\mathbf{L}$. All video KV-Caches are stored for future retrieval. For extremely long videos, we manage memory constraints by offloading KV-Caches to RAM or disk, as in~\citep{infllm}.

\noindent\textbf{External Video KV-Cache Retrieval.} 
Here, we utilize an external CLIP-like model~\citep{clip, siglip} to retrieve question-relevant video KV-Cache, primarily as a baseline to assess whether retrieval can enhance VideoQA performance, as demonstrated in Section~\ref{sec:experiments}. 
Specifically, a CLIP-like model transformers each video frame into a vector $\mathbf{v} = f_v(v) \in \mathrm{R}^D$, where $f_v$ represents the visual encoder, $D$ denotes the vector dimension. Similarly, the question is encoded as $\mathbf{q} = f_t(q) \in \mathrm{R}^D$, where $f_t$ is the text encoder. We then compute the cosine similarity between the embeddings of frame and question: 
\begin{equation}
    \text{Sim}(\mathbf{v}, \mathbf{q}) = \frac{\mathbf{v} \cdot \mathbf{q}}{\tau~||\mathbf{v}||~||\mathbf{q}||}
    \label{equ:retrieval}
\end{equation} 
where $\tau$ is a learnable temperature parameter. This similarity is calculated at the frame level, rather than at the token level. Alternatively, we can group $b$ consecutive frames into blocks by averaging their frame vectors and then compute block-level similarity scores. Finally, the $r$ most relevant video frames or $\lceil r/b \rceil$ video blocks are retrieved. The corresponding video KV-Cache, denoted as $\mathbf{R}$, is subsequently loaded onto the GPU for question-answering.

\noindent\textbf{Internal Video KV-Cache Retrieval.} Building on recent advancements in handling long sequences with LLMs~\citep{infllm, quickllama, em_llm}, we further explore using self-attention layers within Video-LLMs for retrieval. Similar to external retrieval, internal retrieval is still performed at the level of video frames or blocks.

During video modeling, the average of the key vectors of a frame is computed as its representative frame vector: $\mathbf{v} = \frac{1}{N_f} \sum_{j=1}^{N_f} \mathbf{k}_j \in \mathrm{R}^{D'}$, where $N_f$ is the number of tokens per frame, and $\mathbf{k}_j$ is the $j$-th key vector. 
To reduce computational overhead, we do not differentiate between attention heads and instead concatenate them into a single vector, with $D'$ as the resultant dimension. Similarly, the question vector is computed as $\mathbf{q} = \frac{1}{N_q} \sum_{k=1}^{N_q} \mathbf{q}_{k} \in \mathrm{R}^{D'}$, where $N_q$ is the number of tokens in the question, 
and $\mathbf{q}_{k}$ is its $k$-th query vector. The similarity computation and video KV-Cache retrieval are identical to that of external retrieval, except that $\tau$ is set to 1.

Note that, internal retrieval offers several advantages over external retrieval. 
First, it operates independently within each self-attention layer, allowing different layers to retrieve different video blocks.\footnote{For simplicity, we omit the layer index in the above explanation.}
This allows for a broader capture of video context. 
Additionally, internal retrieval reuses already computed hidden representations and does not introduce extra parameters, which reduces the computational overhead compared to external retrieval.

\noindent\textbf{Question-Answering Using Retrieved KV.} The retrieved Video KV-Caches serve as the context for video question-answering. Formally, the attention calculation is formulated as:
\begin{equation}
    \mathbf{O} = \text{Attn}\left(\mathbf{W_Q}\mathbf{X}, [\mathbf{R}_k, \mathbf{W_K}\mathbf{X}], [\mathbf{R}_v, \mathbf{W_V}\mathbf{X}]\right),
    \label{equ:answer_generation}
\end{equation} 
where $\mathbf{X}$ represents either the question tokens or the current token being decoded, and $\mathbf{R}_k$ and $\mathbf{R}_v$ are the key and value vectors from the context, which includes the retrieved video, question, and previously generated tokens. 

\noindent\textbf{Positional Encoding.}
Our baseline Video-LLMs employ Rotary Position Embeddings (RoPE)~\citep{su2024roformer}, a commonly used relative positional encoding method.
Our video streaming encoding process follows LM-Infinite~\citep{lm_infinite}, where RoPE operates normally within the local window but is constrained by a ``distance ceiling'' for more distant tokens. 
For question-answering, we do not account for the original positions of the retrieved KV-Caches, as handling unseen distances among tokens presents significant challenges~\citep{lm_infinite}.
Instead, we treat these retrieved tokens as regular consecutive tokens. 
We also experimented with a static variation from Inf-LLM~\citep{infllm}, where all retrieved tokens are assigned the same position. 
Our results show that applying standard RoPE to retrieved video tokens leads to better performance, likely due to the importance of capturing temporal information in video comprehension.

\section{Experiments}
\label{sec:experiments}

\subsection{Benchmark and Metrics}

\noindent\textbf{MLVU}$_\texttt{dev-mc}$~\citep{mlvu} is the multiple-choice subset of the MLVU-dev benchmark. It focuses on evaluating the long-form video understanding of MLLMs. The question-answer pairs are manually labeled and can be divided into 3 groups: single-detail, multi-detail, and holistic. The evaluation metric is Accuracy.

\noindent\textbf{\textsc{QaEgo4D}}$_\texttt{test-mc}$~\citep{di2023groundvqa} is the multiple-choice subset of the \textsc{QaEgo4D}-test benchmark, focusing on question-answering in long egocentric videos. It includes annotations marking video segments relevant to each question. The evaluation metric is Accuracy.

\noindent\textbf{EgoSchema}~\citep{egoschema} is a diagnostic benchmark for long VideoQA, featuring over 5000 multiple-choice questions and long temporal certificate length. It challenges AI models with long-term understanding, as current state-of-the-art models achieve significantly lower accuracy compared to human performance.

\noindent\textbf{ActivityNet-QA}~\citep{activitynet_qa} 
encompasses human-annotated QA pairs on 5,800 videos derived from the ActivityNet~\citep{activitynet} dataset. 
This benchmark is designed to assess the capabilities of VideoQA models in long-term spatiotemporal reasoning. Our evaluation methodology aligns with that of Video-ChatGPT~\citep{video_chatgpt}, employing \texttt{GPT-3.5-turbo-0613} to judge the accuracy of the open-ended VideoQA responses.

\begin{wraptable}{r}{0.54\textwidth}
\setlength{\tabcolsep}{1.5mm}
\centering
\footnotesize
\vspace{-12pt}
\caption{
\textbf{Summary of the evaluation benchmarks.} 
MC stands for multiple-choice VideoQA, while OE refers to open-ended VideoQA.
}
\vspace{-5pt}
\label{tab:benchmark_comparison}
\resizebox{\linewidth}{!}{
\begin{tabular}{lrrrr}
    \toprule
    Benchmark & Duration & \#Videos & \#QA & Type \\
    \midrule
    MLVU$_\texttt{dev-mc}$ & 12 min & 1,242 & 2,175 & MC \\
    \textsc{QaEgo4D}$_\texttt{test-mc}$ & 8.3 min & 148 & 500 & MC \\
    EgoSchema & 3 min & 5,031 & 5,031 & MC \\
    ActivityNet-QA & 2 min & 800 & 8,000 & OE \\
    RVS-Ego & 60 min & 10 & 1,465 & OE \\
    RVS-Movie & 30 min & 22 & 1,905 & OE \\
    CGBench$_\texttt{mc}$ & 27 min & 1,219 & 12,129 & MC \\
    \bottomrule
\vspace{-20pt}
\end{tabular}
}
\end{wraptable}

\noindent\textbf{RVS-Ego} and \textbf{RVS-Movie}~\citep{flashvstream} are Streaming VideoQA benchmarks, constructed using 10 long videos from the Ego4D dataset~\citep{ego4d} and 22 long videos from the MovieNet dataset~\citep{movienet}, respectively. 
These benchmarks feature open-ended questions paired with timestamps, 
which are initially generated by GPT-4V~\citep{gpt4v} and GPT-4~\citep{gpt4}, and subsequently refined through manual filtering.

\noindent\textbf{CGBench}$_\texttt{mc}$~\citep{cgbench}, the multiple-choice subset of CGBench, is designed for clue-grounded question answering in long videos. It focuses on the ability to retrieve relevant clues for questions, making it an ideal testbed for~\ourmethod.

\subsection{Implementation Details}
We primarily evaluate our approach by integrating it into \texttt{LLaVA-OV-0.5B} and \texttt{LLaVA-OV-7B}~\citep{llava_onevision}, chosen for their simplicity and strong performance.
In the Appendix, we conduct experiments with several other Video-LLMs as further validations.

All experiments are conducted on NVIDIA A100 (80GB) GPUs with FP16 precision.
For video modeling, we process the video stream at 0.5 FPS, in line with GPT-4o's testing on MLVU~\citep{mlvu}. 
The local window size is set to 15K.
For external video KV-Cache retrieval, we use \texttt{SigLIP-SO400M}~\citep{siglip} as the retriever.
For internal KV-Cache retrieval, we set the block size ($b$) to 1 and the number of retrieved frames ($r$) to 64 by default, with further hyper-parameter variations explored in Section~\ref{sec:exp_ablations}.

Unless otherwise specified, \textbf{\ourmethod} refers to the use of internal video KV-Cache retrieval.

\subsection{Ablations}
\label{sec:exp_ablations}
In this section, we conduct ablation studies on the effectiveness of in-context retrieval, number of retrieved frames, and the block size.

\begin{wraptable}{r}{0.48\textwidth}
\centering
\vspace{-17pt}
\caption{
\textbf{Ablation study on \textsc{QaEgo4D}$_\texttt{test-mc}$}.
\colorbox[RGB]{235,235,235}{``Oracle Retrieval''} refers to a scenario where the annotated, question-relevant video segments are used as input, with a uniform sampling of up to 16 frames.
This setup, by definition, has 100\% recall and defines the upper-bound VideoQA performance.
}
\label{tab:ablation_qaego4d}
\vspace{-4pt}
\resizebox{\linewidth}{!}{
\begin{tabular}{l cc}
    \toprule
    Retrieval Method    & VideoQA Acc. & Recall \\
    \midrule
    \footnotesize{\texttt{\textcolor{gray}{LLaVA-OV-0.5B}}} \\
    Uniform Sampling    & 42.6    & 6.1 \\
    External Retrieval  & 48.0    & 58.1 \\
    Internal Retrieval  & 50.0    & 63.4 \\
    \rowcolor[RGB]{235,235,235}
    Oracle Retrieval    & 52.0    & 100 \\
    \midrule
    \footnotesize{\texttt{\textcolor{gray}{LLaVA-OV-7B}}} \\
    Uniform Sampling    & 53.0    & 6.1 \\
    External Retrieval  & 54.2    & 58.1 \\
    Internal Retrieval  & 56.0    & 70.5\\
    \rowcolor[RGB]{235,235,235}
    Oracle Retrieval    & 64.4    & 100 \\
    \bottomrule
\end{tabular}
}
\vspace{-15pt}
\end{wraptable}

\textbf{Effectiveness of In-context Retrieval.}
The experiments on \textsc{QaEgo4D}$_{\texttt{test-mc}}$, as presented in Table~\ref{tab:ablation_qaego4d}, demonstrate the effects of various retrieval methods on VideoQA accuracy and recall.
The recall metric, defined as the percentage of question-relevant video frames retrieved, exhibits a strong positive correlation with VideoQA performance: higher recall consistently leads to better accuracy.
Uniform Sampling, which sparsely selects frames, achieves the lowest recall and, consequently, the poorest VideoQA accuracy. 
In contrast, Oracle Retrieval, with perfect recall, delivers the highest VideoQA accuracy, significantly outperforming Uniform Sampling.
While External and Internal Retrieval fall short of Oracle-level precision, both surpass Uniform Sampling, with Internal Retrieval excelling due to its higher recall. 

The MLVU benchmark~\citep{mlvu} encompasses three types of VideoQA tasks: \textit{Single Detail} requires identifying a single critical plot within a long video, \textit{Multi Detail} necessitates the integration of multiple plots, and \textit{Holistic} demands a comprehensive understanding of the entire video.
This makes MLVU an ideal platform for evaluating our in-context retrieval method. 
As shown in Table~\ref{tab:ablation_mlvu}, both External and Internal Retrieval enhance the overall VideoQA accuracy over the Uniform Sampling baseline. 
The enhancements are most pronounced in Single Detail tasks, 
demonstrating that \ourmethod~effectively retrieves question-relevant video context. Furthermore, Internal Retrieval significantly outperforms External Retrieval in Holistic tasks, likely due to its ability to capture broader context and leverage the Video-LLM’s video modeling capabilities, as discussed in Section~\ref{sec:method}.

\begin{figure}[t]
\setlength{\belowcaptionskip}{0.5\baselineskip}
\centering
    \begin{subfigure}{.5\linewidth}
        \centering
        \includegraphics[width=.95\linewidth]{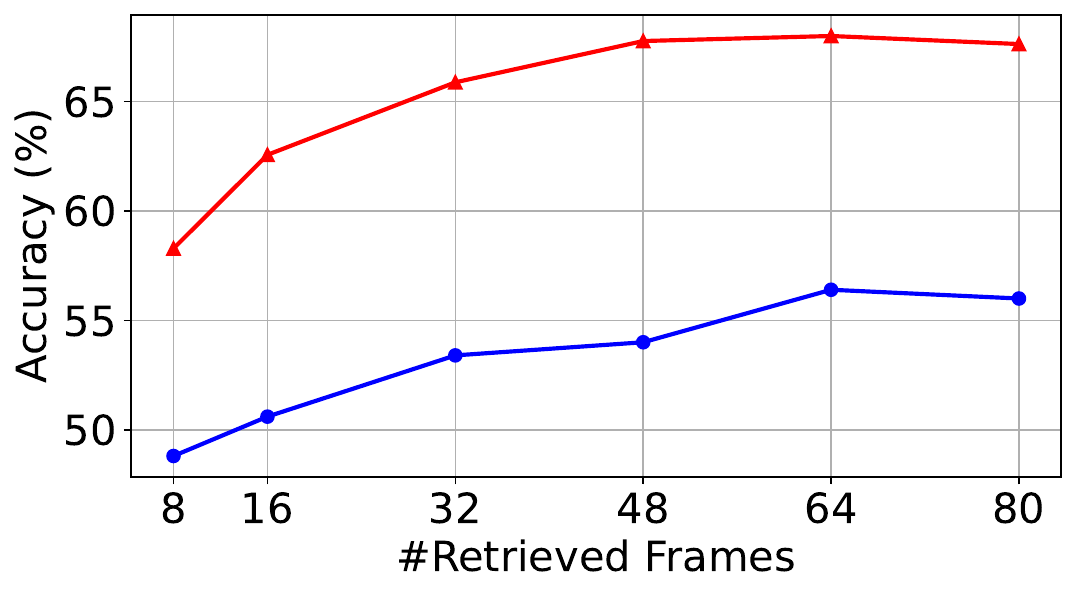}
        \caption{}
    \end{subfigure}%
    \hfill
    \begin{subfigure}{.5\linewidth}
        \centering
        \includegraphics[width=.95\linewidth]{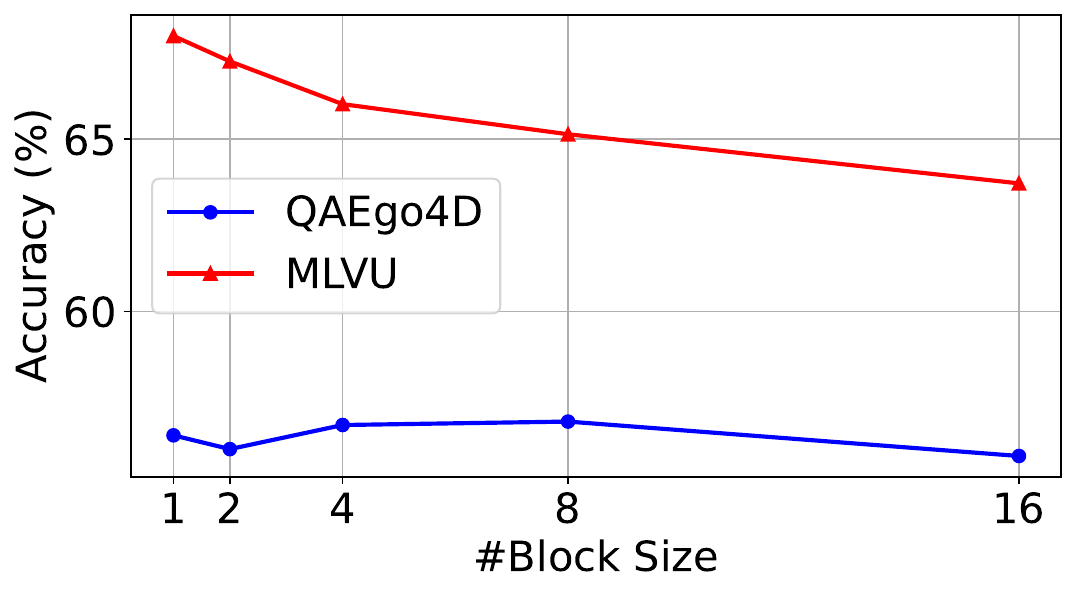}
        \caption{}
    \end{subfigure}
\vspace{-15pt}
\caption{
\textbf{Ablation study of retrieval hyperparameters:}
(a) number of retrieved frames and (b) number of frames per retrieval block.
Experiments are conducted with \texttt{LLaVA-OV-7B}.
}
\label{fig:ablation_retrieval}
\end{figure}

\textbf{Number of Retrieved Frames.}
We fix the block size ($b = 1$) and evaluate the impact of varying the numbers of retrieved frames ($r \in \{8, 16, 32, 48, 64, 80\}$) on the \textsc{QaEgo4D} and MLVU benchmarks.
As illustrated in Figure~\ref{fig:ablation_retrieval}(a), increasing the number of retrieved frames generally improves VideoQA accuracy, as it implies capturing more relevant visual context. 
However, on MLVU, this improvement plateaus as more frames are retrieved since the additional irrelevant information hinders the subsequent question-answering process. Additionally, retrieving more frames increases the computational overhead of the question-answering stage, 
further slowing down inference.

\textbf{Retrieval Block Size.}
When processing video streams, we group $b$ consecutive frames into blocks for block-level retrieval. For this experiment, we fix the number of retrieved frames at $r = 64$ and evaluate different block sizes ($b \in {1, 2, 4, 8, 16}$).
With a fixed $r$, larger block sizes result in fewer, more concentrated retrieved blocks.
Figure~\ref{fig:ablation_retrieval}(b) shows that increasing block size negatively affects accuracy on MLVU, while performance on \textsc{QaEgo4D} remains relatively stable.
This suggests that MLVU tasks benefit from retrieving more dispersed visual cues, aligning with its design of multi-detail and holistic tasks~\citep{mlvu}. 
In contrast, \textsc{QaEgo4D} primarily relies on a single relevant clip per question~\citep{di2023groundvqa}.

\begin{table}[t]
\setlength{\tabcolsep}{1.8mm}
\centering
\caption{
\textbf{Ablation study on MLVU$_\texttt{dev-mc}$.}
 The experiments are based on \texttt{LLaVA-OV-7B}.
}
\label{tab:ablation_mlvu}
\footnotesize
\resizebox{\linewidth}{!}{
\begin{tabular}{l c ccc c cc c cc c c}
    \toprule
    \multirow{2}{*}{Task} &\phantom{a}& \multicolumn{3}{c}{Single Detail} &\phantom{a}& \multicolumn{2}{c}{Multi Detail} &\phantom{a}& \multicolumn{2}{c}{Holistic} &\phantom{a}& \multirow{2}{*}{Avg.} \\
    \cmidrule{3-5} \cmidrule{7-8} \cmidrule{10-11}
                       && Needle & Ego  & PlotQA && Order & Count && Topic & Anomaly &&  \\
    \midrule
    Uniform Sampling    && 74.1   & 59.7 & 69.8   && 45.9  & 32.0  && 87.9  & 72.0    && 64.7 \\
    External Retrieval  && 78.6   & 69.6 & 71.6   && 40.2  & 37.9  && 84.5  & 63.0    && 66.3 \\
    Internal Retrieval  && 75.8   & 66.6 & 76.3   && 45.2  & 36.9  && 90.1  & 74.5    && 68.5 \\
    \bottomrule
\end{tabular}
}
\end{table}

\subsection{Offline Video Question-answering}

\begin{table}[t]
\setlength{\tabcolsep}{1.5mm}
\centering
\caption{
\textbf{Offline video question-answering on four long-form benchmarks.}
``Acc.'' denotes accuracy, and
``Score'' is the open-ended answer rating by \texttt{gpt-3.5-turbo-0613} on a scale from 1 to 5.
}
\label{tab:sota}
\footnotesize
\resizebox{\linewidth}{!}{
\begin{tabu}{l l l l ll}
    \toprule
    \multirow{2}{*}{Method} & MLVU & \textsc{QaEgo4D} & EgoSchema  & \multicolumn{2}{l}{ActivityNet-QA} \\
    \cmidrule{2-6}
     & dev Acc. & test Acc. & Acc. & Acc. & Score \\
    \midrule   
    \rowcolor[RGB]{235,235,235}
    GPT-4V~\citep{gpt4v}             & 49.2 & - & -    & 57.0 & -    \\
    \rowcolor[RGB]{235,235,235}
    GPT-4o~\citep{gpt4o}          
    & 64.6 & - & -    & -    & -    \\
    \rowcolor[RGB]{235,235,235}
    Gemini-1.5-Flash~\citep{gemini}  & -    & - & 65.7 & 55.3 & -    \\
    \rowcolor[RGB]{235,235,235}
    Gemini-1.5-Pro~\citep{gemini}    & -    & - & 72.2 & 57.5 & -    \\
    \hdashline
    
    Video-ChatGPT-7B~\citep{video_chatgpt} & 31.3 & -    & -    & - & - \\
    LLaMA-VID-7B~\citep{llamavid}    & 33.2 & -    & -    & 47.4 & 3.30 \\
    MiniGPT4-Video-7B~\citep{minigpt4_video} & 44.5 & - & - & 44.3 & 3.35 \\
    Video-LLaVA-7B~\citep{video_llava}   & 47.3 & -    & -    & - & - \\
    LongVA-7B~\citep{longva}         & 56.3 & - & -    & 50.0 & -    \\
    VideoStreaming~\citep{videostreaming} & - & - & 44.1 & - & - \\
    Flash-VStream-7B~\citep{flashvstream} & 50.2 & 38.2 & 38.1 & 51.9 & 3.40 \\
    \hdashline
    
    LLaVA-OV-0.5B~\citep{llava_onevision}     & 53.2 & 42.6 & 29.6 & 50.5 & 3.02 \\
    \quad\textbf{+\ourmethod} \scriptsize(\retrieve{0.5}{64})              & \textbf{56.1}\plusvalue{2.9} & \textbf{50.0}\plusvalue{7.4} & \textbf{31.0}\plusvalue{1.4} & \textbf{52.1}\plusvalue{1.6} & \textbf{3.15}\plusvalue{.13} \\
    LLaVA-OV-7B~\citep{llava_onevision}       & 64.7 & 52.8 & 59.8 & 56.6 & 3.29 \\
    \quad\textbf{+\ourmethod} \scriptsize(\retrieve{0.5}{64})              & \textbf{68.5}\plusvalue{3.8} & \textbf{56.0}\plusvalue{3.2} & \textbf{60.7}\plusvalue{0.9} & \textbf{60.4}\plusvalue{3.8} & \textbf{3.52}\plusvalue{.23} \\
    \bottomrule
\end{tabu}
}
\end{table}

Streaming video understanding is a relatively under-explored area, with limited StreamingVQA benchmarks available~\citep{flashvstream}.
As discussed in Section~\ref{sec:task},
OfflineVQA can be considered as a special case of StreamingVQA.
Thus, we first evaluate our method in the offline setting using four widely adopted long-form VideoQA benchmarks, comparing our results against state-of-the-art VideoQA methods.
A summary of these benchmarks can be found in Table~\ref{tab:benchmark_comparison}.

As shown in Table~\ref{tab:sota}, our proposed \ourmethod~always enhances the performance of \texttt{LLaVA-OV-0.5B} and \texttt{LLaVA-OV-7B} without additional training.
Notably, \texttt{LLaVA-OV-7B}+\ourmethod~outperforms two memory-based StreamingVQA models (VideoStreaming~\citep{videostreaming} and Flash-VStream~\citep{flashvstream}) by a large margin. 
While the base model already demonstrates strong performance, 
and we \textbf{do not} claim credit for this achievement, 
our method can integrate seamlessly with Video-LLMs, benefiting from their ongoing advancements.

\subsection{Streaming Video Question-answering} \label{sec:streaming}

\begin{table}[t]
\setlength{\tabcolsep}{1.8mm}
\centering
\caption{
\textbf{StreamingVQA benchmark results.}
All methods are tested under identical conditions.
``Video Enc.'' is frames encoded per second. 
``Latency'' is measured from question input to response completion.
``GPU'' indicates peak GPU memory usage, and ``KV-Cache'' refers to the video KV-Cache size offloaded per hour.
}
\label{tab:streamingvqa}
\resizebox{\linewidth}{!}{
\begin{tabular}{p{3cm}c ccc ccc ccc cc}
    \toprule
    \multirow{2}{*}{Retrieval Method} && \multicolumn{2}{c}{RVS-Ego} && \multicolumn{2}{c}{RVS-Movie} && \multicolumn{2}{c}{Running Speed} && \multicolumn{2}{c}{Memory Usage} \\
    \cmidrule{3-4} \cmidrule{6-7} \cmidrule{9-10} \cmidrule{12-13}
                                && Acc. & Score && Acc. & Score && Video Enc. & Latency && GPU & KV-Cache \\
    \midrule
    Flash-VStream-7B && 57.3  & 4.0  && 53.1  & 3.3  && 14 FPS & 2.4s && 20 GB & - \\
    \hdashline
    \footnotesize{\texttt{\textcolor{gray}{LLaVA-OV-7B}}} \\
    Uniform Sampling            && 56.2  & 3.7  && 43.0  & 3.3  &&  -        & 2.9s && 21 GB & - \\
    External Retrieval && 62.4  & 3.9  && 53.6  & 3.5  &&  11 FPS   & 5.8s && 55 GB & 18.8 GB/h \\
    Internal Retrieval && 63.7  & 4.0  && 54.4  & 3.6  &&  11 FPS   & 3.3s && 38 GB & 18.8 GB/h \\
    \hdashline
    \footnotesize{\texttt{\textcolor{gray}{LLaVA-OV-0.5B}}} \\
    Uniform Sampling            && 51.8  & 3.7  && 37.2  & 3.2  &&  -        & 2.5s && 7 GB   & - \\
    External Retrieval && 54.1  & 3.8  && 44.7  & 3.4  &&  17 FPS   & 4.1s && 37 GB  & 4.0 GB/h \\
    Internal Retrieval && 54.7  & 3.9  && 44.6  & 3.4  &&  17 FPS   & 1.6s && 19 GB  & 4.0 GB/h \\
    \bottomrule
\end{tabular}
}
\end{table}

We then evaluate our method on the streaming setting using the RVS-Ego and RVS-Movie benchmarks. During video stream modeling, questions are input immediately after their annotated end timestamps and answered based on the preceding video content.

\noindent\textbf{Question-answering Performance.}
Table~\ref{tab:streamingvqa} presents the StreamingVQA performance. 
Both external and internal retrieval methods significantly outperform the uniform sampling baseline. 
Additionally, our approach enables \texttt{LLaVA-OV-7B} to surpass Flash-VStream~\citep{flashvstream}, demonstrating \ourmethod's effectiveness for the StreamingVQA.

\textbf{Running Speed and Memory Usage.}
We also examine the running speed and memory usage under controlled conditions. 
Specifically, a 1-hour, 1080P video from RVS-Ego with 100 scattered questions is used. 
Each question is padded to 64 tokens, and the generated answers are fixed at 128 tokens in length. 
The video frames are pre-extracted at 0.5 FPS (1,800 frames in total) and streamed to the Video-LLM frame by frame. 

As illustrated in Table~\ref{tab:streamingvqa}, both retrieval methods maintain high video encoding speeds, with \texttt{LLaVA-OV-7B} achieving 11 FPS and \texttt{LLaVA-OV-0.5B} achieving 17 FPS.
Moreover, KV-Cache offloading remains manageable, with \texttt{LLaVA-OV-7B} at 18.8GB/h and \texttt{LLaVA-OV-0.5B} at 4.0GB/h (see appendix for more details).
External retrieval, however, introduces higher latency and GPU memory usage due to additional computations in the external retriever, whereas internal retrieval significantly reduces both.
Figure~\ref{fig:teaser} has also demonstrated that latency and GPU memory usage remain stable as more frames are processed.
Flash-VStream also shows good efficiency. However, it only maintains a relatively small memory footprint (681 tokens)~\citep{flashvstream}, leading to potential information loss when dealing with extremely long videos.

\begin{figure}[t]
\centerline{\includegraphics[width=\linewidth]{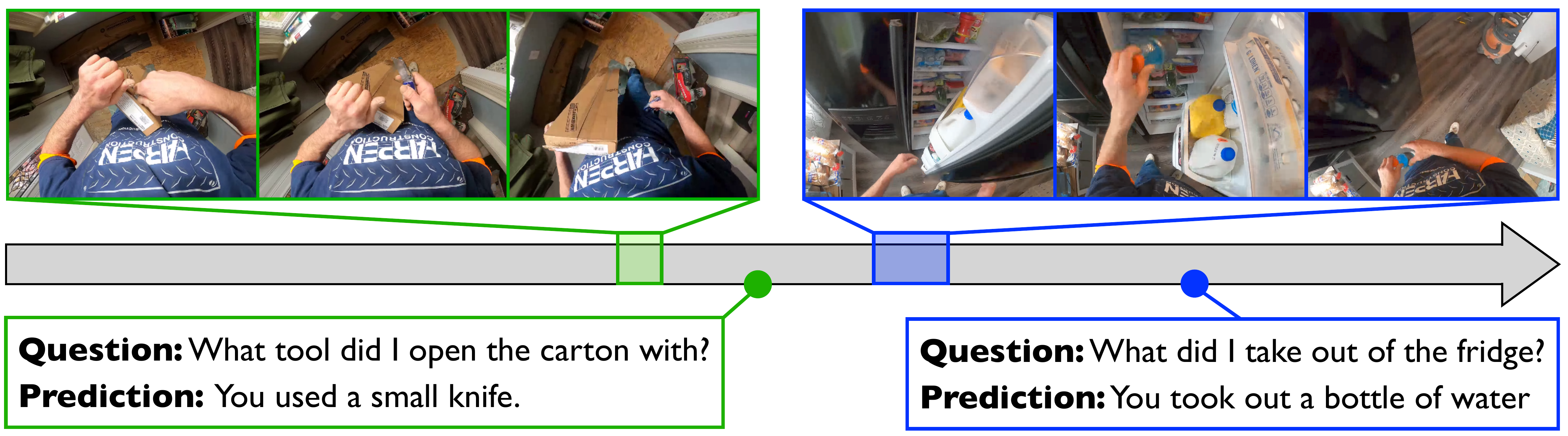}}
\caption{
\textbf{StreamingVQA qualitative examples.}
The example is drawn from the \textsc{QaEgo4D} benchmark.
The video stream is processed frame by frame.
\textcolor[RGB]{35,166,8}{$\CIRCLE$} and \textcolor{blue}{$\CIRCLE$} mark the timestamps at which questions are posed.
\textcolor[RGB]{35,166,8}{$\square$} and \textcolor{blue}{$\square$} indicate the relevant video contexts that support answering these questions.
}
\label{fig:visualization}
\end{figure}

\textbf{Qualitative Examples.}
Figure~\ref{fig:visualization} presents an example of streaming video question-answering.
Our approach continuously processes video streams while responding to questions posed at different timestamps.
To improve efficiency, it stores and retrieves relevant video KV-Caches as contextual information for answering these questions.

We provide additional implementation details and experimental results in the Appendix.

\section{Related Work}
\label{sec:related_work}

\noindent\textbf{LLMs for Video Understanding.} 
In recent years, there has been a surge of interest in leveraging Large Language Models (LLMs) for video understanding, leading to the development of several innovative approaches~\citep{video_chatgpt, llava_next_video, llava_onevision}. These models typically use a Vision Encoder to extract video features, followed by a mapping step with Linear Projection, MLP, or Q-Former~\citep{blip2}. The mapped features are combined with textual data and fed into large language models (LLMs) to generate a text output. These models have relatively simple architectures, requiring less training data and computational resources, yet they achieve strong performance on short video understanding benchmarks~\citep{msrvtt_qa, next_qa, li2024mvbench}. However, they employ sparse sampling or token compression techniques to reduce the number of tokens, which can result in significant information loss when dealing with longer or more content-rich videos. As a result, they are not well-suited for long video understanding or streaming video understanding.

\noindent\textbf{Long Video Understanding.} 
A central challenge in long video understanding is effectively compressing the information from lengthy videos. Many approaches use language as a bridge, condensing videos into dense captions~\citep{llovi,video_recap,streaming_dvc}. While this achieves good results in some cases, compressing video content into text often leads to the loss of crucial visual details. Besides, as a pioneering approach to streaming video understanding, VideoLLM-Online~\citep{videollm_online} employs a data-centric methodology by interleaving video and text during training. In contrast, our approach is training-free, allowing seamless integration with various existing Video-LLMs to extend their StreamingVQA capabilities. Additionally, VideoLLM-Online retains only a single token per frame to handle long videos, which may result in visual information loss. Our method preserves complete visual information and leverages In-Context KV-Cache Retrieval to enhance efficiency.

Another line of research focuses on compressing long videos into a memory bank~\citep{wu2019long, wu2022memvit, wang2023memory}.
MC-ViT~\citep{mc_vit} adapts existing pretrained video transformers by fine-tuning them to attend to condensed visual memories. It relates closely to the token-pruning, merging, and memory-based video understanding methods. In comparison, we propose a training-free method specifically tailored to the StreamingVQA task. Incorporating MC-ViT into the StreamingVQA task could be an interesting avenue for future research, and we acknowledge its potential in this domain.
This approach has been integrated into Video-LLMs for streaming video understanding, as shown in works like VideoStreaming~\citep{videostreaming} and Flash-VStream~\citep{flashvstream}. These methods dynamically update the memory during video processing and utilize it for downstream tasks.
Despite their innovation, a major limitation of these methods is their failure to account for video length and information density, especially when using a fixed memory size. For example, Flash-VStream compresses both 10-second clips and hour-long videos into the same 681 tokens.
Furthermore, these methods lack interpretability, making it difficult to determine how much information is being compressed into the memory or whether relevant video information is being accurately retrieved during downstream tasks.

In pursuit of greater interpretability in long video understanding, methods such as GroundVQA~\citep{di2023groundvqa} and GeLM~\citep{chen2024groundedmultihopvideoqalongform} advocate for localizing relevant video clips while responding to user queries. Drawing inspiration from these, this work refrains from excessively condensing video information. By harnessing the causal capabilities of Video-LLMs, it preserves the entire Video KV-Cache, allowing for the retrieval of relevant information when required. This strategy effectively mitigates the substantial loss of video content while improving interpretability.

\noindent\textbf{Long Context Handling for LLMs.} 
Handling long text sequences in LLMs has been a major challenge due to high computational and memory costs, leading to training constraints on shorter sequences.
Techniques like StreamingLLM~\citep{streamingllm} and LM-Infinite~\citep{lm_infinite} use sliding window attention to process long sequences incrementally, but discard distant tokens, limiting the model’s ability to capture long-range dependencies. Recent approaches~\citep{infllm,quickllama,em_llm} address this by storing and retrieving previously computed KV-Caches, enabling better recall of distant contexts.

\noindent\textbf{Retrieval-Augmented Generation.} 
Retrieval-augmented generation (RAG) combines retrieval mechanisms with generative models to enhance performance across various NLP tasks by incorporating external knowledge~\citep{guu2020retrieval,lewis2020retrieval,borgeaud2022improving} and improving performance in vision-language tasks~\citep{xu2024retrieval}. In-context retrieval, recently proposed for handling long inputs~\citep{ram2023context}, retrieves information from the input document itself rather than an external knowledge base. In-context KV-Cache retrieval further improves efficiency by pre-encoding long documents, avoiding redundant encodings, and leveraging the LLM’s own retrieval capabilities for faster, more effective performance.
\section{Conclusion}
\label{sec:conclusions}

In conclusion, this paper introduces a training-free approach, \ourmethod, designed to enhance the efficiency of Video Large Language Models (Video-LLMs) for streaming video question-answering (StreamingVQA). 
Unlike conventional video question-answering (VideoQA) systems that must process entire videos before answering, \ourmethod~enables rapid, real-time responses. 
By employing a sliding-window attention mechanism, it ensures that the model only considers a subset of previous frames while encoding the video stream, significantly cutting down on computational demands. To retain key video context, we developed an in-context KV-Cache retrieval method that efficiently stores and reloads key-value vectors that relevant for each query. This targeted retrieval strategy, combined with the ability to perform video modeling and question-answering on separate processes and GPUs, results in a highly efficient streaming VideoQA system. Extensive experiments show that \ourmethod~not only surpasses existing VideoQA models in performance but also enhances their practicality for real-world streaming applications.

\vspace{5pt}

\textbf{Acknowledgements.}
This work is supported by National Key R\&D Program of China (No. 2022ZD0161400).
We thank Yikun Liu for discussions and conducting experiments on CGBench.

\clearpage
\bibliography{iclr2025_conference}
\bibliographystyle{iclr2025_conference}

\clearpage
\lstset{
basicstyle=\small\ttfamily,
columns=flexible,
breaklines=true,
moredelim=**[is][\color{red}]{@}{@}
}

\appendix

In the appendix, we provide additional implementation details, experiments, and discussions of limitations and future work.

\section{Additional Implementation Details}

\subsection{Multi-processing Serving}

As discussed in Section~\ref{sec:task}, our approach enables the separation of video modeling and question-answering across different processes and GPUs, significantly enhancing efficiency in real-world applications. Specifically, we dedicate a primary process for video stream encoding, utilizing sliding-window attention to analyze the video and store the computed cache in RAM. If RAM capacity is exceeded, the data can be offloaded to disk. Additionally, a process pool is maintained, with the number of processes determined by the frequency of queries and available resources. Each process loads the same Video-LLM parameters but operates independently. The video processing continues uninterrupted, without waiting for question-answering tasks to complete. When a query is posed, we log its timestamp to ensure that video information after this point is excluded from the answer. An available process from the pool is then activated to retrieve relevant video key-value vectors using our method, loading them onto its GPU for question-answering. This approach enables efficient StreamingVQA applications, with significant potential in areas such as robotics, surveillance, augmented reality, and live broadcasting.

\subsection{Prompt Templates for VideoQA}

We use the same prompt template for all multiple-choice VideoQA benchmarks. Text in~\texttt{\textcolor{red}{red}} indicates variable inputs.

\begin{lstlisting}
System:
You are a helpful assistant.
User: 
@<video>@
Question: @<question>@
Options:
(A) @<Option_A>@
(B) @<Option_B>@
(C) @<Option_C>@
(D) @<Option_D>@
(E) @<Option_E>@
Answer with the option's letter from the given choices directly.
Assistant:
\end{lstlisting}

The prompt template for open-ended VideoQA is rather simpler:

\begin{lstlisting}
System:
You are a helpful assistant.
User: 
@<video>@
@<question>@
Assistant:
\end{lstlisting}

\subsection{KV-Cache Size Calculation}

The size of the KV-Cache can be calculated using the following formula, assuming FP16 precision:
\begin{equation*}
    2 \times L~\text{layers} \times T~\text{frames} \times M~\text{tokens/frame} \times H~\text{heads} \times D~\text{dimension} \times 2~\text{bytes}.
\end{equation*}

For \texttt{LLaVA-OV-7B}~\citep{llava_onevision}, with $L=28$, $M=196$, $H=4$, and $D=128$, processing a 1-hour video at 0.5 FPS ($T=1800$) results in a total KV-Cache size of 18.8 GB.

Similarly, for \texttt{LLaVA-OV-0.5B}~\citep{llava_onevision}, with $L=24$, $M=196$, $H=2$, and $D=64$, processing a 1-hour video at 0.5 FPS results in a total KV-Cache size of 4.0 GB.

These theoretical calculations are consistent with the experimental results shown in Table~\ref{tab:streamingvqa}.

\section{Additional Experiments}

\subsection{Experiments with more Video-LLMs and Benchmark}

To further assess the generalizability of our approach, we tested it on additional Video-LLMs: \texttt{Video-LLaVA-7B}~\citep{video_llava}, \texttt{LongVA-7B}~\citep{longva}, and \texttt{LLaVA-OV-72B}~\citep{llava_onevision}. We used model sharding for \texttt{LLaVA-OV-72B}, significantly slowing inference. To mitigate this, we reduced the FPS to 0.1 and the number of retrieved frames to 32, ensuring efficient evaluation. As shown in \Cref{tab:additional}, ReKV consistently improved performance across various models and benchmarks, highlighting its robustness and adaptability.

\begin{table}[h]
\centering
\caption{
\textbf{Additional experiments with more Video-LLMs and benchmark.}\\
``Acc.'' denotes accuracy.
``X Frames'' refers to uniformly sampling X frames from the video.
``Y FPS $\to$ X Frames'' indicates an input video with a frame rate of Y FPS, from which X frames are retrieved.
}
\label{tab:additional}
\vspace{-5pt}
\footnotesize
\resizebox{\linewidth}{!}{
\begin{tabu}{l l l l l l}
    \toprule
    \multirow{2}{*}{Method} & \multirow{2}{*}{Sampling} & MLVU & \textsc{QaEgo4D} & EgoSchema  & CGBench \\
    \cmidrule{3-6}
    & & dev Acc. & test Acc. & Acc. & Acc. \\
    \midrule   
    Video-LLaVA-7B~\citep{video_llava} & 8 Frames     & 46.5 & 37.0 & 41.4 & 18.7 \\
    \quad\textbf{+\ourmethod} & \retrieve{0.5}{8}             & \textbf{47.2}\plusvalue{0.7} & \textbf{37.9}\plusvalue{0.9} & \textbf{42.2}\plusvalue{0.8} & \textbf{19.2}\plusvalue{0.5} \\
    \hdashline
    LongVA-7B~\citep{longva} & 32 Frames    & 57.3 & 42.8 & 42.5 & 26.1 \\    
    \quad\textbf{+\ourmethod} & \retrieve{0.5}{32}             & \textbf{58.6}\plusvalue{1.3} & \textbf{45.6}\plusvalue{2.8} & \textbf{42.7}\plusvalue{0.2} & \textbf{26.4}\plusvalue{0.3} \\

    \hdashline
    
    LLaVA-OV-0.5B~\citep{llava_onevision} & 64 Frames    & 53.2 & 42.6 & 29.6 & 21.4 \\
    \quad\textbf{+\ourmethod} & \retrieve{0.5}{64}              & \textbf{56.1}\plusvalue{2.9} & \textbf{50.0}\plusvalue{7.4} & \textbf{31.0}\plusvalue{1.4} & \textbf{21.7}\plusvalue{0.3} \\
    \hdashline
    LLaVA-OV-7B~\citep{llava_onevision} & 64 Frames      & 64.7 & 52.8 & 59.8 & 31.1 \\
    \quad\textbf{+\ourmethod} & \retrieve{0.5}{64}              & \textbf{68.5}\plusvalue{3.8} & \textbf{56.0}\plusvalue{3.2} & \textbf{60.7}\plusvalue{0.9} & \textbf{33.9}\plusvalue{2.8} \\
    \hdashline
    LLaVA-OV-72B~\citep{llava_onevision} & 32 Frames      & 71.9 & 53.6 & 59.6 & 37.2 \\
    \quad\textbf{+\ourmethod} & \retrieve{0.1}{32}              & \textbf{72.6}\plusvalue{0.7} & \textbf{57.0}\plusvalue{3.4} & \textbf{62.0}\plusvalue{2.4} & \textbf{40.5}\plusvalue{3.3} \\
    \bottomrule
\end{tabu}
}
\end{table}

\subsection{Fair comparisons with Flash-VStream} \label{sec:fair_comparison}

\Cref{tab:sota,tab:streamingvqa} compared \texttt{LLaVA-OneVision+ReKV} with \texttt{Flash-VStream}. However, these comparisons may be unfair due to different architecture and training data. Thus, here we conduct \textbf{fair} comparisons using the same Video-LLM backbone, including the identical visual encoder (\texttt{CLIP-ViT-L/14}), projector (2-layer MLP), LLM (\texttt{Vicuna-7B-v1.5}), training data, and train/eval pipelines.

Due to the inaccessibility of WebVid videos\footnote{https://github.com/m-bain/webvid} used in Flash-VStream’s original training, we use 232K randomly sampled InternVid videos\footnote{https://huggingface.co/datasets/OpenGVLab/InternVid} as a substitute. This ensures comparable experimental settings.
We train a baseline Video-LLM model (\texttt{Base}) and a Flash-VStream-enhanced version (\texttt{Base+Flash}). Similarly, we integrate ReKV into the same baseline (\texttt{Base+ReKV}) for fair comparisons.
To maintain parity, the baseline uniformly samples 16 frames per video, resized to $224\times224$. Visual features of shape $(T, 16, 16, D)$ are average-pooled to $(T, 8, 8, D)$ before being passed through the MLP projector and into the LLM.
Both Flash-VStream and ReKV process video at 0.5 FPS, with ReKV retrieving 16 frames.

\begin{table}[h]
\centering
\caption{
\textbf{Fair comparisons with Flash-VStream.}
\textcolor{gray}{``Original Flash''} is the checkpoint officially published by Flash-VStream while ``Base+Flash'' is our reproduced version.
}
\label{tab:flash}
\vspace{-5pt}
\footnotesize
\resizebox{0.9\linewidth}{!}{
\begin{tabu}{l l l l l l}
    \toprule
    Method & MLVU$_\texttt{dev-mc}$ & \textsc{QaEgo4D}$_\texttt{test-mc}$ & EgoSchema  & RVS-Movie & RVS-Ego \\
    \midrule   
    Base & 49.8 & 39.0 & 42.6 & 47.2 & 54.1 \\
    Base+Flash & 51.0	& 37.4	& 41.2	& 50.1	& \textbf{55.4} \\
    \textbf{Base+ReKV}	& \textbf{51.9}\plusvalue{0.9}	& \textbf{40.5}\plusvalue{3.1}	& \textbf{43.7}\plusvalue{2.5}	& \textbf{51.9}\plusvalue{1.8}	& 54.7\minusvalue{0.7} \\
    \rowfont{\color{gray}} Original Flash	& 50.2	& 38.2	& 38.1	& 53.1	& 57.3 \\
    \bottomrule
\end{tabu}
}
\end{table}

As shown in Table~\ref{tab:flash}, \texttt{Base+ReKV} consistently outperforms the base Video-LLM \texttt{Base} and surpasses \texttt{Base+Flash} in most cases, highlighting its superiority under fair comparative conditions.
Additionally, ReKV offers enhanced usability, seamlessly integrating with existing Video-LLMs without requiring extensive retraining.

On the contrary, the reproduced \texttt{Base+Flash} does not consistently outperform \texttt{Base}. It excels on StreamingVQA (RVS-Movie and RVS-Ego) and MLVU but underperforms on \textsc{QAEgo4D} and EgoSchema.
This discrepancy is likely due to significant visual information loss: the \texttt{Base} model processes 1024 visual tokens ($16\times64$), while \texttt{Base+Flash} uses only 681 memory tokens.

For additional context, we include results from the original Flash-VStream (\texttt{Original Flash}) using checkpoints from its official repository\footnote{https://github.com/IVGSZ/Flash-VStream}.
Our reproduced \texttt{Base+Flash} shows performance deviations, likely due to differences in training data and potential environmental factors.

\subsection{Computational Complexity}

We ensure \textbf{fair} comparisons by using the identical Video-LLM backbone (Sec.~\ref{sec:fair_comparison}) under controlled streaming conditions (Sec.~\ref{sec:streaming}). Specifically, we measured the FLOPs and MACs of the base Video-LLM, Flash-VStream, and our external and internal retrieval methods. We analyzed \textbf{average TFLOPs and TMACs per QA over various question frequencies} in a 1-hour video, leveraging the \texttt{calflops} library~\citep{ye2023calflops}.

As shown in~\Cref{tab:tflops,tab:tmacs}, ReKV’s efficiency improves significantly with increasing QA frequency.
The video stream is encoded only once, and computed results are reused across QAs, leading to reduced per-query complexity as QA frequency rises.
Flash-VStream outperforms ReKV at low QA frequencies (\textit{e.g.}, 100 QAs). However, ReKV’s complexity decreases more rapidly with increased QA frequency, primarily due to Flash-VStream’s high memory update overhead.
ReKV is thus better suited for high-concurrency scenarios such as live streaming and requires no additional training.

Furthermore, Internal retrieval consistently outperforms external retrieval, reducing average FLOPs by 15.5\% and MACs by 15.2\%.
These results underscore ReKV’s ability to balance computational efficiency and effectiveness, particularly in dynamic, high-query environments.
This positions ReKV as a practical and scalable solution for streaming video understanding.

\begin{table}[h]
\setlength{\tabcolsep}{1.5mm}
\centering
\caption{
\textbf{TFLOPs / QA.}
}
\label{tab:tflops}
\vspace{-5pt}
\footnotesize
\resizebox{0.7\linewidth}{!}{
\begin{tabu}{l cccc}
    \toprule
    \#QAs & Baseline &Flash-VStream & ReKV (External) & ReKV (Internal) \\
    \midrule   
    100 &	22.4 &	\textbf{15.5} &	21.7 &	18.5 \\
    200 &	12.7 &	14.1 &	11.4 &	\textbf{9.6} \\
    360 &	8.5 &	13.8 &	6.8 &	\textbf{5.6} \\
    \bottomrule
\end{tabu}
}
\vspace{-5pt}
\end{table}

\begin{table}[h]
\setlength{\tabcolsep}{1.5mm}
\centering
\caption{
\textbf{TMACs / QA.}
}
\label{tab:tmacs}
\vspace{-5pt}
\footnotesize
\resizebox{0.7\linewidth}{!}{
\begin{tabu}{l cccc}
    \toprule
    \#QAs & Baseline &Flash-VStream & ReKV (External) & ReKV (Internal) \\
    \midrule   
    100 &	11.2 &	\textbf{7.8} &	10.8 &	9.2 \\
    200 &	6.4 &	7.1 &	5.7 &	\textbf{4.8} \\
    360 &	4.3 &	6.8 &	3.3 &	\textbf{2.8} \\
    \bottomrule
\end{tabu}
}
\vspace{-5pt}
\end{table}

\section{Limitations and Future Work}
\label{sec:limitation}

While \ourmethod~improves the accuracy and efficiency of Video-LLMs in the StreamingVQA task, it still has several limitations that deserves future investigation:
{\em First}, although the KV-Cache offloading to RAM or disk is manageable, as shown in Table~\ref{tab:streamingvqa}, handling extremely long video streams, such as those in surveillance, may lead to an unsustainable increase in cache size. This issue can be mitigated by integrating techniques such as quantization, token pruning, and compression.
{\em Second}, the use of a constant block size for grouping consecutive frames during retrieval can disrupt video continuity. A more refined solution would involve segmenting videos into semantically coherent blocks.
{\em Third}, our method retrieves a fixed number of frames. Future work could explore dynamic retrieval strategies that adjust the number of frames based on video context and query requirements.
{\em Finally}, StreamingVQA remains an under-explored task with few available benchmarks. Developing high-quality benchmarks with precise temporal annotations is crucial for advancing future research.

\end{document}